\documentclass[conference]{IEEEtran}
\IEEEoverridecommandlockouts
\usepackage{amsmath,amssymb,amsfonts}
\usepackage[hidelinks]{hyperref}
\usepackage{algorithmic}
\usepackage{graphicx}
\usepackage{textcomp}
\usepackage{xcolor}
\def\BibTeX{{\rm B\kern-.05em{\sc i\kern-.025em b}\kern-.08em
    T\kern-.1667em\lower.7ex\hbox{E}\kern-.125emX}}
    
\usepackage[style=ieee]{biblatex}
\graphicspath{{img}}

\usepackage{stfloats}
\usepackage[abbreviations]{glossaries-extra}

\glssetcategoryattribute{acronym}{indexonlyfirst}{true}

\newabbreviation{ft}{FT}{Fourier Transform}
\newabbreviation{stft}{STFT}{Short-Time Fourier Transform}
\newabbreviation{simo}{SIMO}{Single Input Multiple Output}
\newabbreviation{cw}{CW}{Continuous-Wave}
\newabbreviation{acma}{ACMA}{Analytical Constant Modulus Algorithm}
\newabbreviation{racma}{RACMA}{Real Analytical Constant Modulus Algorithm}
\newabbreviation{ica}{ICA}{Independent Component Analysis}
\newabbreviation{emd}{EMD}{Empirical Mode Decomposition}
\newabbreviation{rbm}{RBM}{Random Body Movement}
\newabbreviation{rsm}{RSM}{Radar Self-Motion}
\newabbreviation{nmf}{NMF}{Non-Negative Matrix Factorization}
\newabbreviation{vae}{VAE}{Variational Autoencoder}
\newabbreviation{wae}{WAE}{Wasserstein Autoencoder}
\newabbreviation{cnn}{CNN}{convolutional neural network}
\newabbreviation{elbo}{ELBO}{Evidence Lower Bound}
\newabbreviation{mmd}{MMD}{Maximum Mean Discrepancy}
\newabbreviation{ot}{OT}{Optimal Transport}
\newabbreviation{ppg}{PPG}{Photoplethysmography}
\newabbreviation{ecg}{ECG}{Electrocardiogram}
\newabbreviation{pcg}{PCG}{Phonocardiogram}
\newabbreviation{sir}{SIR}{Signal-to-Interference Ratio}
\newabbreviation{kld}{KLD}{Kullback-Leibler Divergence}
    
\begin{document}

\title{Interference Motion Removal for Doppler Radar Vital Sign Detection Using Variational Encoder-Decoder Neural Network
% {\footnotesize \textsuperscript{*}Note: Sub-titles are not captured in Xplore and
% should not be used}
\thanks{This work was funded by EPSRC under Grant EP/R513349/1.}
}

\author{\IEEEauthorblockN{
Mikolaj~Czerkawski,
Christos~Ilioudis,
Carmine~Clemente,
Craig~Michie,
Ivan~Andonovic,
Christos~Tachtatzis
}
                                    % ...
%\\

\IEEEauthorblockA{% 1st affiliations
Department of Electronic and Electrical Engineering, University of Strathclyde, Glasgow, G1~1XW, UK}
%\IEEEauthorblockA{\IEEEauthorrefmark{2}% 2nd affiliations
%(Affiliation): dept. name of organization, name/acronyms of organization, City, Country, e-mail address*}
%\IEEEauthorblockA{\IEEEauthorrefmark{3}% 3rd affiliations
%(Affiliation): dept. name of organization, name/acronyms of organization, City, Country, e-mail address*}
%\IEEEauthorblockA{\IEEEauthorrefmark{4}% 4th affiliations
%(Affiliation): dept. name of organization, name/acronyms of organization, City, Country,
% e-mail address*} 
}

\maketitle

\begin{abstract}
    The treatment of interfering motion contributions remains one of the key challenges in the domain of radar-based vital sign monitoring. Removal of the interference to extract the vital sign contributions is demanding due to overlapping Doppler bands, the complex structure of the interference motions and significant variations in the power levels of their contributions. A novel approach to the removal of interference through the use of a probabilistic deep learning model is presented. Results show that a convolutional encoder-decoder neural network with a variational objective is capable of learning a meaningful representation space of vital sign Doppler-time distribution facilitating their extraction from a mixture signal. The approach is tested on semi-experimental data containing real vital sign signatures and simulated returns from interfering body motions. The application of the proposed network enhances the extraction of the micro-Doppler frequency corresponding to the respiration rate is demonstrated.
\end{abstract}


\section{Introduction}

    % Describe the general problem (vs radar monitoring)
    The application of radar for vital sign detection has been subject to extensive research over the last decade~\cite{Alizadeh2019, Yang2019} owing to the many advantages inherent within non-contact health monitoring. A radar-based system has the potential for the continuous and unobtrusive tracking of respiration rate and heart rate of one or multiple living targets with no invasion. A solution requiring physical contact is also undesirable in a variety of applications, such as patients with compromised skin or infant monitoring.
    
    % Describe specific problem (rbm interference)
    A number of core challenges in the context of the system design can be identified; here, the problem of vital sign extraction in the presence of interfering returns from additional body motions (e.g. limbs) is addressed. The extraction of vital sign signals using radar relies on mechanical oscillations of the body surface induced by displacements owing to both respiration and heart beat. These motions can be extracted from the Doppler shifts contained in a received radar signal, particularly difficult in the presence of other non-stationary reflections in the field of view. Further, interference may not necessarily be confined to separate objects as it may also originate from the other parts of the monitored target. The limbs of an ostensibly stationary living target still move in a fairly random fashion, or perform approximately static activities such as moving their arm and fingers while using a mobile phone; this class of interference motions is generally described as \gls*{rbm}.
    
    % Related Work
    %\subsection{Related Work}
    % A - VITAL SIGNS & RBM
    One of the earliest reports on the blind separation of human vital signs in Doppler radar signals was presented in~\cite{Petrochilos2007}. A \gls*{simo} \gls*{cw} radar system with one transmitter and two receivers was utilised to separate multiple contributions in the signal with \gls*{acma}, \gls*{racma} and \gls*{ica} algorithms explored. \gls*{emd} was applied in~ \cite{Mostafanezhad2013} to the received single-channel bandpass-filtered signal. In~\cite{Dirksmeyer2019} an analytically derived method for respiration signal separation was presented; however, the method is not appropriate for signals with noise. In~\cite{Ye2020} and~\cite{Ye2020-2} a similar blind source separation approach is applied with different clustering methods relying on \gls*{nmf}. In~\cite{Wang2019} an unsupervised machine learning framework is presented and, similarly to~\cite{Ye2020, Ye2020-2} rely on the sparsity of the cardiac signal in time. In~\cite{Gu2019} a deep neural network is used to process an unwrapped phase signal and remove the interference of body movement. A review of methods for \gls*{rbm} cancellation (as well as \gls*{rsm}) is provided in~\cite{Zhu2019}, ranging from the use of multiple radar transceivers~\cite{Li2008, Wang2011}, through a camera-assisted radar~\cite{Gu2013} to a matched filter approach.~\cite{Lv2018} However, the majority of the approaches rely on the application of multiple sensors, increasing the complexity and the practical implementation of the systems significantly.

    % B - BLIND SOURCE SEPARATION
    A body of research has been published in the recent years on the application of \gls*{vae} architectures to source separation, predominantly in the audio domain. In~\cite{Pandey2018} a separate \gls*{vae} model is trained for each expected source within the mixed signal in order to extract each contribution from a magnitude of \gls*{stft}. Both~\cite{Kameoka2019} and~\cite{Seki2019} operate on complex \gls*{stft} outputs instead, addressing the challenge of determined and under-determined separation, respectively. In the latter, a generalised version of the multi-channel \gls*{vae} approach is introduced. A somewhat similar problem of audio source separation with weak class supervision is addressed in~\cite{Karamatli2019} using the architecture from~\cite{Hsu2017}, where the first layer of the convolutional network has kernels that extend over an entire (yet single) instantaneous spectrum.
    
    The contribution presented in this work constitutes the first application of a variational encoder-decoder neural network for interference removal from vital signs radar returns. The interference removal framework operates solely on single channel complex radar signals and the network operation can be flexibly adjusted by controlling the datasets used for training. This means that the same method can be applied for a wide range of other applications where certain contributions in a radar signal are to be removed. Furthermore, the probabilistic nature of the architecture allows the model to learn a more realistic representations of the signal, where a given input is matched with a distribution of possible satisfactory output samples rather than a single sample.
    
    Section~\ref{sec:problem_formulation} contains the problem formulation, which is followed by Section~\ref{sec:proposed_solution} that describes the proposed solution. The results and a demonstration of micro-Doppler based respiration rate extraction are presented in Section~\ref{sec:results}. Section~\ref{sec:conclusion} contains concluding remarks.

\section{Problem Formulation}
\label{sec:problem_formulation}
    
        % \subsection{Radar Signals}
        % 1. What are vital sign contributions like?
        Respiration and heart beat result in mechanical oscillations of the body surface in several locations of the human body. Since both contributions are generally expected to be present on the chest, this is the area where a radar device is directed in the reported experiments. The respiration component is not only of higher amplitude but also present over a larger area, each contribution is modelled as separate scattering point $s_b(t)$, similarly to~\cite{Czerkawski2020}. Both $s_b(t)$ and $s_h(t)$ are described in a similar manner, with the range traces $R_b(t)$ and $R_h(t)$ corresponding to the motion associated with the respiration and the heart beat, respectively. For a carrier wavelength of $\lambda$ and effective component magnitudes $|X_b|$ and $|X_h|$ the two components are described as:
        \begin{equation}
        \label{eq:breath_intro}
        s_b(t) = |X_b|e^{j4\pi\frac{R_b(t)}{\lambda}}
        \end{equation}
        \begin{equation}
        \label{eq:heart_intro}
        s_h(t) = |X_h|e^{j4\pi\frac{R_h(t)}{\lambda}}
        \end{equation}
        
        % 2. Nature of Random Body Motions
        Random body movement is most often composed of several moving parts. However, since the velocity trace of each reflecting point can be approximated by a mono-component signal, all random body movement $i(t)$ can be treated as a sum of $N$ reflecting components with different strengths $|X_k|$ and varying range traces $R_k(t)$:
        \begin{equation}
        \label{eq:rbm_intro}
        i(t) = \sum_{k=1}^{N}|X_k|e^{j4\pi\frac{R_k(t)}{\lambda}}
        \end{equation}
        
        % 3. Complete Contribution
        Finally, the total contribution $\rho(t)$ from a human target can be described as a sum of the vital sign contributions and the random body movements:
        
        \begin{equation}
        \label{eq:total_intro}
        \rho(t) = |X_b|e^{j4\pi\frac{R_b(t)}{\lambda}} + |X_h|e^{j4\pi\frac{R_h(t)}{\lambda}} + \sum_{k=1}^{N}|X_k|e^{j4\pi\frac{R_k(t)}{\lambda}}
        \end{equation}
        
        The range or velocity traces can be extracted from the radar range-time and velocity-time maps respectively. In the case of non-modulated \gls*{cw} radar, only velocity-time maps can be derived. The vital signs contributions and the \gls*{rbm} are expected to have overlapping Doppler bands and significant differences in their power levels. Additionally, the \gls*{rbm} and other sources of interference can have fairly complex structure making them difficult to model using non-stochastic processing.

    \begin{figure}[tb]
        \centering
        \includegraphics[width=\columnwidth]{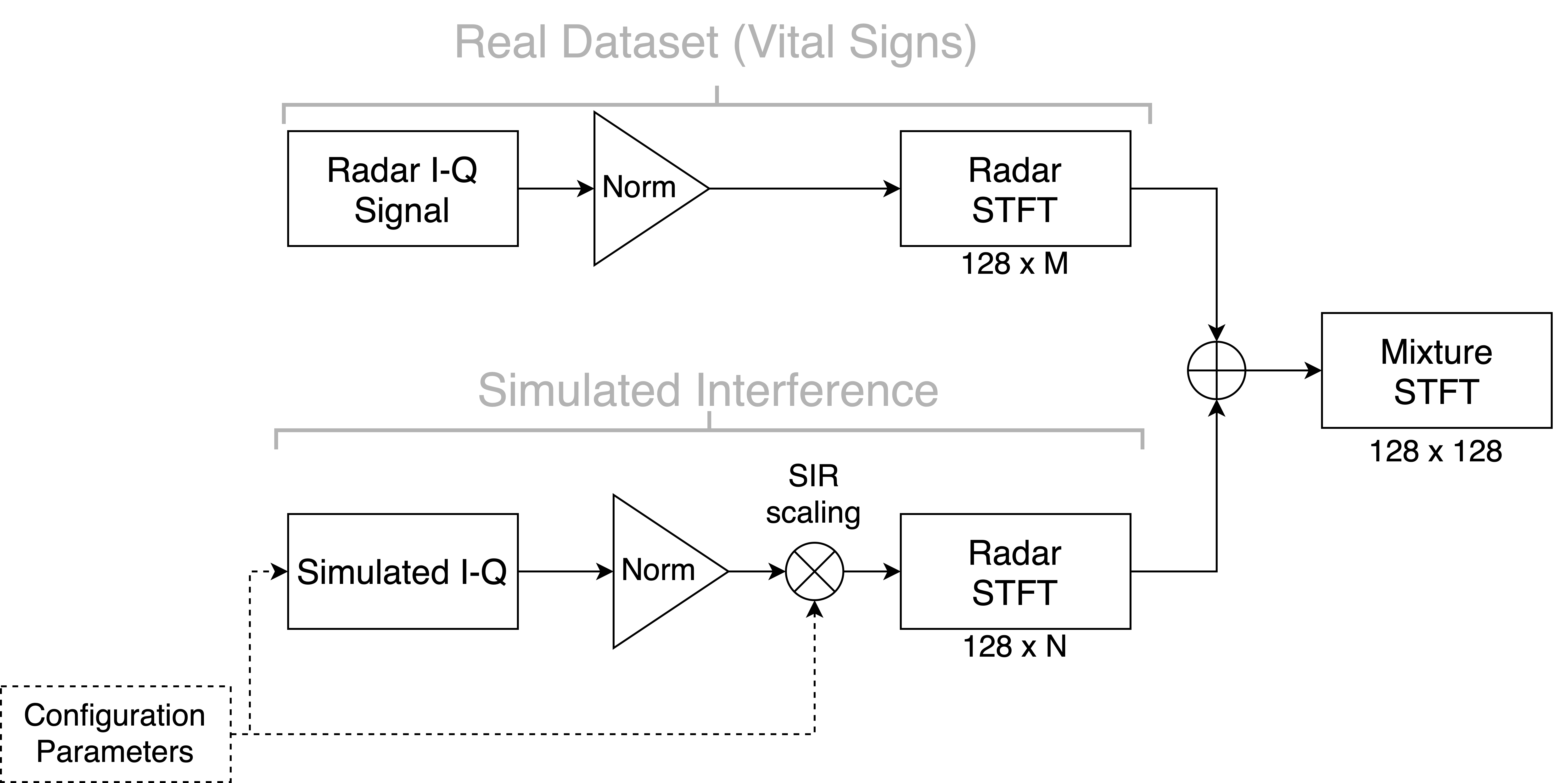}
        \caption{Sample synthesis pipeline}
        \label{fig:pipeline}
    \end{figure} 

\section{Proposed Solution}
\label{sec:proposed_solution}
% Questions:
% 1. Why not separation in 1D? (Doppler spectrum is an important feature, transformation to Doppler domain is treated as a prior)
% 2. 1D may make it harder to separate a larger number of reflecting components (to detect multiple velocities simulatneously an abstraction of T-F processing needs to be learned as an extra challenge)
% 3. Why not standard Convolutional AE?
% 4. Why not conditional GAN?

    % Explain the high-level idea and processing pipeline
    The proposed solution utilises a variational \gls*{cnn} as an interference removal system in a fashion analogous to \gls*{vae}. The system is trained and tested in a semi-experimental environment, where the vital sign \gls*{stft}s are derived from real recordings and the interference motion is synthesized according to a widely applied model~\cite{chen2011micro}.
    % Explain where data comes from
    %   a. dataset
    \subsection{Vital Signs Dataset}
    \label{subsec:vs_dataset}
        The real vital signs samples are from an open dataset of ``GUARDIAN Vital Sign Data''~\cite{shi_schellenberger_2019}. The dataset contains a number of reference signals, such as \gls*{pcg}, \gls*{ecg} and respiration sensor in addition to the recording of radar signals\cite{Shi2020}. The radar dataset is not balanced as it contains samples from varying scenarios, sampling rates and sensor placement configurations at different proportions. In order to render the dataset compatible with the needs of the development of a machine learning solution, a balanced subset of the data has been selected, confined to samples with both the \gls*{pcg} sensor and the radar positioned directly in the front of the target. The selection is further restricted to samples contained directly in the corresponding directory and ignores the sub-folders containing more unusual scenarios (such as apnea, post-exercise, speech, or angle variation). A relatively low target sampling rate of 100~Hz is specified and each source signal is decimated to match that rate (this includes a 30-point FIR low-pass filter of degree 16 with Hamming window).
        
        Radar data within the dataset is supplied in the quadrature format, with separate traces for $I$ and $Q$ signals. The power ratio between this signal and the simulated interference motion is controlled by first normalising power of both signals to achieve zero-mean and unit-variance followed by scaling of the interference signal to match a given \gls*{sir} ratio.
        
        The temporal quadrature signal needs to be transformed into a two-dimensional time-frequency distribution. Here, \gls*{stft} is applied and the output distribution can be interpreted as the levels of energy reflected at given Doppler frequencies for a range of time instants. A Blackman window of 200~ms (with 120~ms overlap) is applied with 128 spectral bins to obtain \gls*{stft} of a whole signal. Then, a segment of 128 instantaneous spectra corresponding to 10.24~seconds is randomly selected. Thus, a complex 128~by~128 \gls*{stft} representation is obtained with the size appropriate for the network input.
        
        \begin{table}[bt]
            \centering
            \caption{Parameters Used for Interference Simulation}
            \begin{tabular}{|l|c|}
                \hline
                Forward Motion & Disabled\\
                Duration & 11\\
                Height & [1.2, 1.8)\\
                Relative Velocity & [0.2, 1.0)\\
                Radar Location &(0,10,0)\\
                $\lambda$ & $\frac{c}{24.17 GHz}$\\
                Range Resolution & 0.01\\
                 \hline
            \end{tabular}
            \label{tab:sim_params}
        \end{table}
        
    \begin{figure*}[tb]
        \centering
        \includegraphics[width = \textwidth]{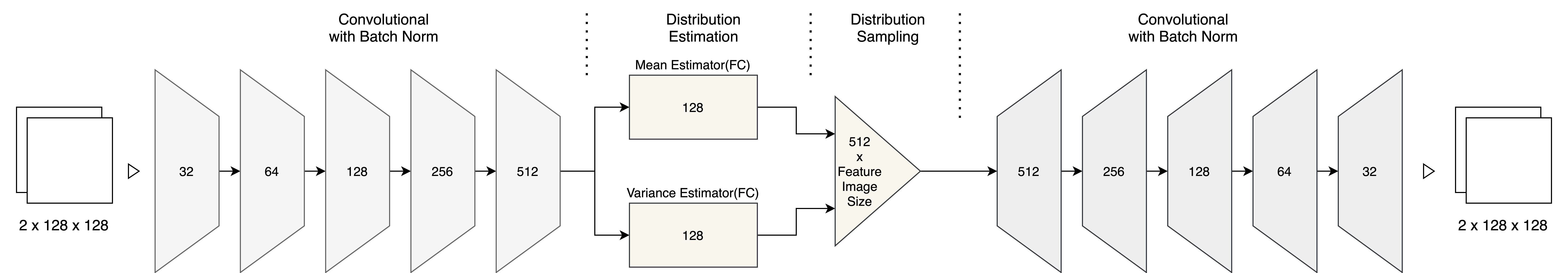}
        \caption{Architecture diagram of the utilised network}
        \label{fig:vae_arch}
    \end{figure*}
        
    %   b. simulator
    \subsection{Human Walking Simulator}
        The simulator is based on a common model derived from experimental bio-mechanical data~\cite{Boulic1990} and the scripts therein are based on the MATLAB scripts written by Chen~\cite{chen2011micro}. A more flexible version of the simulator\footnote{\url{https://doi.org/10.5281/zenodo.4245158}} has been implemented in Python allowing for direct calls and integration within the PyTorch framework. The simulator allows the specification of the configuration file selecting the limbs to be used for synthesis, in turn allowing for a more versatile generation of signals, for example, the swinging of the arms only. Additionally, a script for the generation of static datasets containing human walking radar returns is also featured.
        
        The Python framework can be configured to synthesise functions with a desired sampling rate and duration of the signal, as well as velocities and heights (or uniform sampling ranges thereof). The simulator produces a range-time map, and since the vital signs data is derived from a \gls*{cw} radar, all of the range bins are integrated, creating a one-dimensional signal. To increase the speed of training, a dataset of 1024 samples is synthesised beforehand. In this work, the dataset contains samples with the only interference being the left and right foot walking motions with no net forward motion of the human target. Simulation parameters are listed in Table~\ref{tab:sim_params}.
        
    %   c. combining two sources
    \subsection{Vital Signs Combined with Interference Motion}
        The body motion radar return is simulated at 100 Hz for 11~seconds to match the length of the vital signs \gls*{stft}. After both one-dimensional signals are normalized to zero-mean and unit-variance of magnitude, as shown in Fig.~\ref{fig:pipeline}, the interference signal is multiplied by a factor yielding a specific value of \gls*{sir}. The signals with adjusted power are then subject to a \gls*{stft} with the same parameters. A random 128-unit long temporal segment is then extracted from each \gls*{stft} and both segments are summed to create a 128 by 128 \gls*{stft} image.
    
    \subsection{Architecture}
    
    The architecture shown in Fig.~\ref{fig:vae_arch} is designed to operate on complex \gls*{stft} input with two channels for the real and imaginary component. This way, both channels contain information about the magnitude and phase of the \gls*{stft}, which has been experimentally found to be beneficial. The encoder consists of 5 convolutional layers with depths of 32, 64, 128, 256, and 512 respectively. The output of the last convolutional layer is flattened and passed through two fully connected variational segments with a width of 128, one for the mean estimation and the other for the variance estimation. The latent code is obtained by sampling from the distribution defined by the output of the two estimation units. The decoder takes in the latent code and reconstructs the output image.

    \begin{figure*}[tb]
        \centering
        \includegraphics[width = \textwidth]{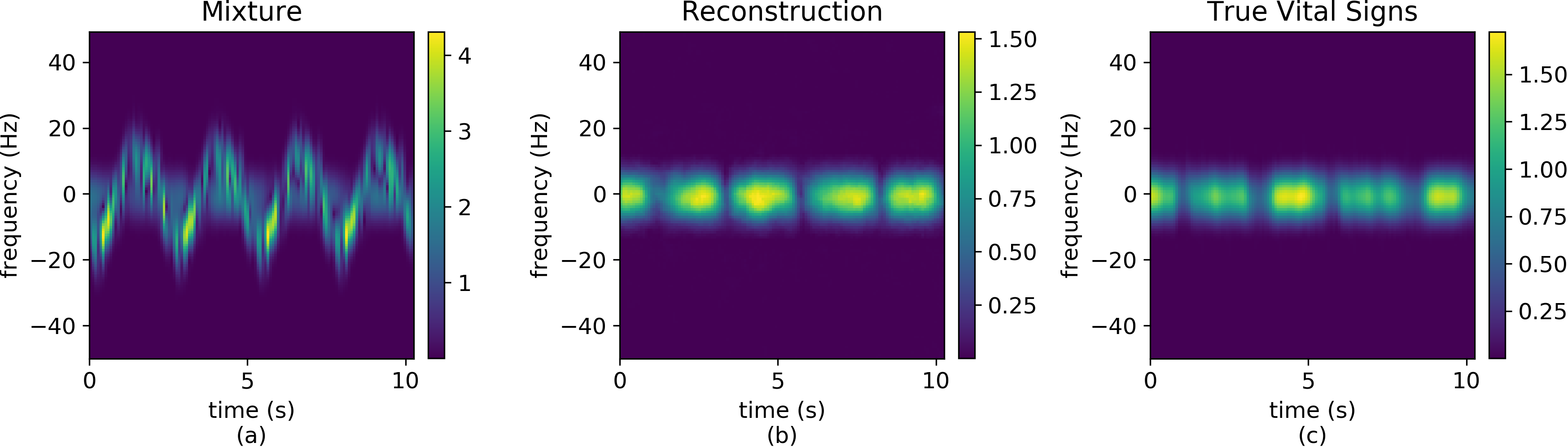}
        \caption{(a) Unseen sample containing vital signs mixed with motion interference (b) Corresponding network output. (c) The ground truth reference. The brightness corresponds to the magnitude of \gls*{stft}.}
        \label{fig:network_output}
    \end{figure*}
    
    \subsection{Training}

        % Training Parameters
        
        \begin{table}[bt]
            \centering
            \caption{Training Parameters}
            \begin{tabular}{|l|l|}
                \hline
                Batch Size & 4 \\
                Optimizer & AdamW\\
                Initial Learning Rate & 1e-3\\
                Scheduler & ReduceLROnPLateau\\
                Scheduler Patience & 32 \\
                \hline
            \end{tabular}
            \label{tab:train_params}
        \end{table}
    
        % Training Stages
        The network is trained in a single stage (details in Table~\ref{tab:train_params}) for 1024 epochs using a sum of two losses - a reconstruction loss and a divergence loss. The weight of the latter has been reduced to $10^{-6}$ from the value suggested in~\cite{Kingma2014} to promote the reconstruction loss during training. The number has been selected experimentally; higher tested values would result in a very low \gls*{kld} loss but the output of the network would be noisy. Checkpointing was applied to restore the weights from epoch 779 that resulted in the lowest validation loss.

\section{Results}
\label{sec:results}

    \begin{figure}[b]
        \centering
        \includegraphics[width = 0.7\columnwidth]{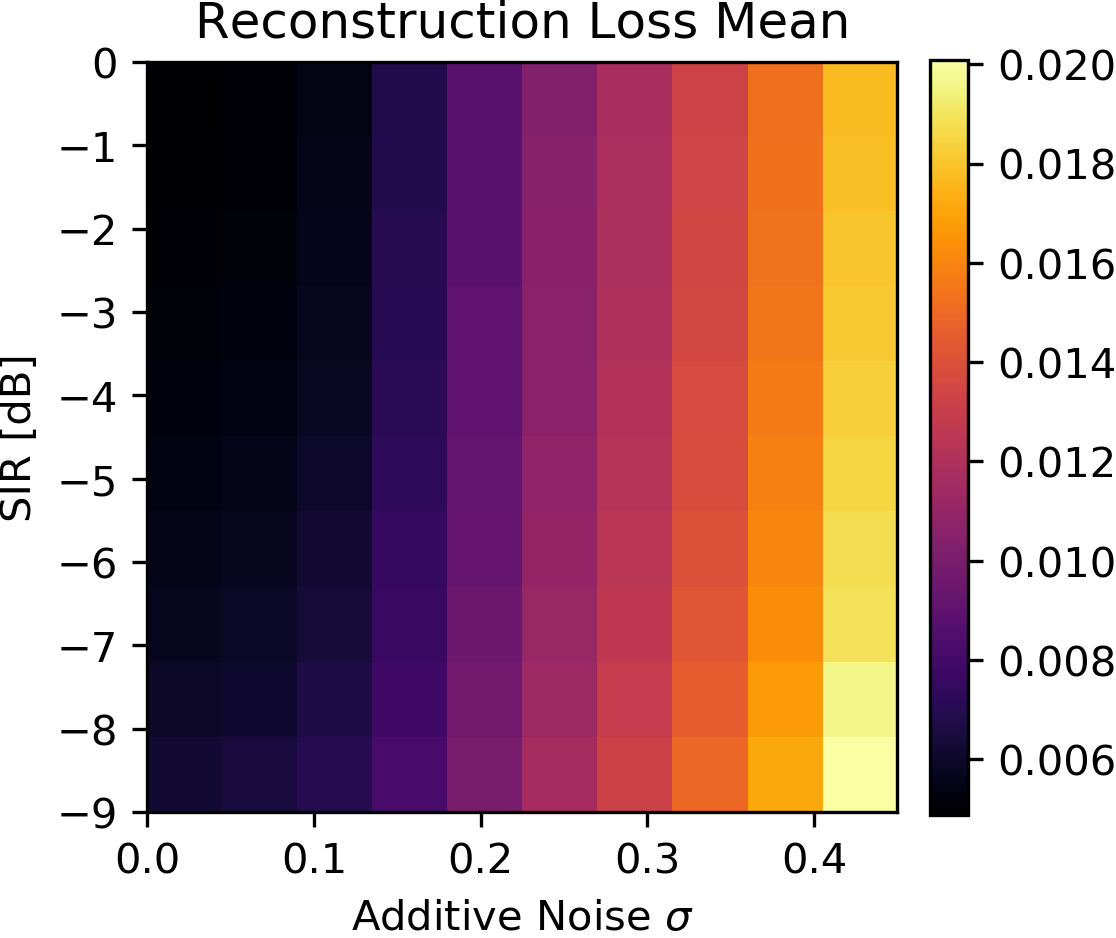}
        \caption{Reconstruction loss influence comparison}
        \label{fig:recons_snr}
    \end{figure}
    % Show basic illustrative results
    The feasibility of the presented approach is demonstrated in two contexts. Firstly, it is shown that the designed network is capable of generalisation beyond the training dataset and can produce outputs with low loss for unseen samples. This intermediate domain is convenient because it is not difficult to compute the loss of the network output and hence provide a direct measure of the system performance. This context is not sufficient in the more general scope where the monitoring of vital signs is of interest. To address that, the second part of the results illustrates how the respiration rate estimation can be carried out using the network output.
    
    % Example of Clean VS & Auto-Encoded Output
    \subsection{Network Operation}
    
    The operation of the network can be illustrated in Fig.~\ref{fig:network_output}. The network takes a complex \gls*{stft} sample in quadrature format containing interference motions (a) and estimates the vital signs contribution to the \gls*{stft} (b). Note that the images shown in Fig.~\ref{fig:network_output} correspond to the magnitude of the complex samples. Some degree of magnitude blurring is apparent in the network output image, however, as it is demonstrated in Section~\ref{subsec:resp_rate_extract}, this does not prevent a successful extraction of the respiration rate.
    
    The performance of the network can be further summarised by the results illustrated in Fig.~\ref{fig:recons_snr}. The network is evaluated by computing a metric of interest for a range of \gls*{sir} values (0 to -9 dB) and for a range of levels of Gaussian noise (standard deviation between 0.0 to 0.45) added to the $I$ and $Q$ channels independently. A direct way of measuring the network performance would be to do this for the mean reconstruction loss. The resulting horizontal gradient observed in Fig.~\ref{fig:recons_snr} demonstrates that the level of added noise is the primary influence over the loss. The minimal differences along the vertical axis suggest that the network can successfully remove the interference at various levels of \gls*{sir} while still producing a reconstruction of equivalent quality.
    
    \begin{figure*}[t]
        \centering
        \includegraphics[width = \textwidth]{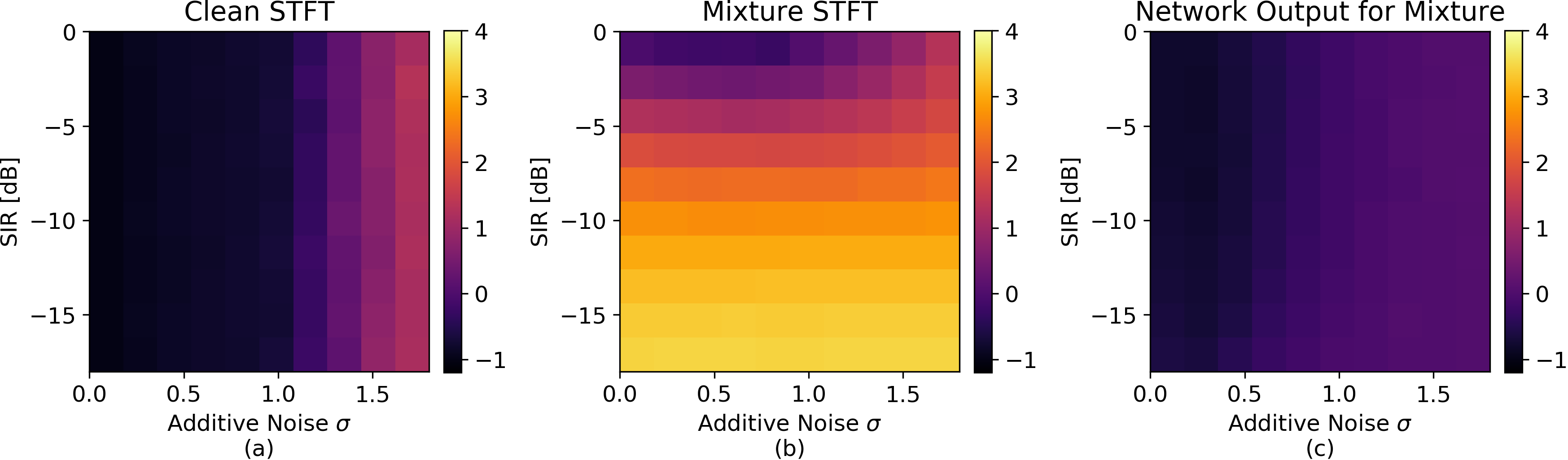}
        \caption{Bin error influence comparison. (a) Clean vital signs sample. (b) Mixture sample. (c) Mixture sample processed by the network. Logarithmic scaling is applied to the displayed error values.}
        \label{fig:sir_snr}
    \end{figure*}
    
    % Proxy Task- Respiration Rate Reconstruction
    \subsection{Respiration Rate Extraction}\label{subsec:resp_rate_extract}
    The utility of using the network for micro-Doppler oriented applications can be assessed by testing the same respiration rate extraction technique for three \gls*{stft} representations: 1)~a clean vital signs sample 2)~sample with added interfering motion contribution 3)~the image of 2) processed by the network. The applied metric is the frequency bin error $\textrm{E}_{b}$ that is defined as the absolute difference between the output of function $\mathfrak{f}(x_{[t]})$ for a reference signal $r_{[t]}$ and for a signal derived from the network output $d_{[t]}$. The network output is a complex matrix and $d_{[t]}$ is obtained by summing all of the Doppler bins.
    
    \begin{equation}
        \textrm{E}_{b}(d_{[t])}, r_{[t])}) = |\mathfrak{f}(d_{[t]}) - \mathfrak{f}(r_{[t]})|
    \end{equation}
    
    For a 1-D signal $x_{[t]}$, the function $\mathfrak{f}(x_{[t]})$ measures the location of the dominant spectrum peak of a temporal signal:
    
    \begin{equation}
        \mathfrak{f}(x_{[t]}) = \left|\mathrm{f}_{DC}-\mathrm{argmax}\left(\left|\mathcal{F}\left\{x_{[t]}\right\}\right|\right)\right|
    \end{equation}
    where $\mathrm{f}_{DC}$ indicates the index of the DC component in the Fourier Transform of the signal $\mathcal{F}\left\{x_{[t]}\right\}$.
    
    Fig.~\ref{fig:sir_snr} illustrates how the mean value of $\textrm{E}_{b}$ changes for three input types (clean vital signs, mixture with interference, and mixture processed by the network) over the same range of \gls*{sir} and additive noise level as in the Fig.~\ref{fig:recons_snr}. The scale and extent are preserved among the three figures, where the brightness of each segment corresponds to the value of mean $\textrm{E}_{b}$ computed for the entire validation dataset. Note that the error values have been converted to logarithm scale in order to make the three graphs easier to compare.
    
    The results for the clean \gls*{stft} could, in fact, be presented in one dimension, since the input remains the same regardless of the \gls*{sir} level. However, due to the probabilistic nature of the network, the output for each sample may vary a little at each iteration of the sweep. This is manifested by minute changes of bin error along the vertical axis. Still, the most significant influence over the clean \gls*{stft} extraction error is the level of added Gaussian noise. The error logarithm stays close to -1 before it begins to increase for the additive noise $\sigma$ values above 1.0.
    
    In contrast, Fig.~\ref{fig:sir_snr}(b) represents the error computed for the same extraction technique with an \gls*{stft} input containing a mixture of the vital signs and the interfering motion. The dominance of the added interference influence is represented by a significant vertical gradient and the highest levels of error (error logarithm near 4.0). A slight increase due to levels of added noise is observed for the highest levels of tested \gls*{sir} but this effect becomes negligible as the \gls*{sir} increases.
    
    Finally, Fig.~\ref{fig:sir_snr}(c) illustrates the error achieved for an input containing the mixed sample processed by the network. The error remains approximately constant for the tested range of \gls*{sir} values. The consistent increase of the error logarithm for higher levels of noise (from -1 to around 0) demonstrates that the network derives the output from the relevant features in the input. Furthermore, the maximum value of mean $\textrm{E}_{b}$ in Fig.~\ref{fig:sir_snr}(c) is significantly lower than in Fig.~\ref{fig:sir_snr}(a) demonstrating the noise suppression property of the network.

\section{Conclusion}
\label{sec:conclusion}
    % 1. Human Simulator Model
    % 2. Novel Application of VAE to Interference Removal
    % 3. Semi-Experimental Demonstration
    The presented contribution demonstrates the first application of a variational encoder-decoder \gls*{cnn} for interference removal from radar returns containing vital signs of human targets. The evaluation has been carried out in a semi-experimental mode, where an open real dataset source has been combined with simulated human walking motions. The realised \gls*{sir} sweeps demonstrate that the network reduces the influence of the interference motion on the vital sign extraction while displaying sensitivity to global additive noise. The latter observation suggests that the network does indeed derive the output from the input features instead of trivialising the task to generation of likely samples. The approach presented in this manuscript could be extended to more challenging interference scenarios and consequently, form a basis for a fully experimental framework.

\printbibliography

\end{document}